\RequirePackage[OT1]{fontenc}
\documentclass[conference]{IEEEtran}
\renewcommand{\IEEEauthorrefmark}[1]{\textsuperscript{#1}}
\usepackage{cite}
\usepackage{graphicx}
\usepackage{amsmath,amsfonts}
\usepackage{booktabs}

\newif\ifdoubleblind
\newcommand{\pmv}[2]{\ensuremath{#1\!\pm\!#2}}

\begin{document}
\bstctlcite{IEEEcontrol}

\title{Predicting Ethanol Drinking Status from Nonhuman Primate Liver Histology: A Proof of Concept for Parameter-efficient Fine-tuning in Few-Shot Vision-Language Modeling}

\ifdoubleblind
\author{\IEEEauthorblockN{Anonymous Authors}
\IEEEauthorblockA{Anonymous Affiliations}}
\else
\author{\IEEEauthorblockN{Amirmohammad Mohammadi\IEEEauthorrefmark{1},
Rupak Khadka\IEEEauthorrefmark{2},
Kathleen A. Grant\IEEEauthorrefmark{2}, \\
Mary Lauren Benton\IEEEauthorrefmark{3}, and
Joshua Peeples\IEEEauthorrefmark{1}}
\IEEEauthorblockA{\IEEEauthorrefmark{1}Department of Electrical and Computer Engineering, Texas A\&M University}
\IEEEauthorblockA{\IEEEauthorrefmark{2}Oregon Health \& Science University}
\IEEEauthorblockA{\IEEEauthorrefmark{3}Department of Computer Science, Baylor University}
\IEEEauthorblockA{Email: \IEEEauthorrefmark{1}\{amir.m, jpeeples\}@tamu.edu,\\
\IEEEauthorrefmark{2}\{khadka, grantka\}@ohsu.edu, \IEEEauthorrefmark{3}marylauren\_benton@baylor.edu}}
\fi

\maketitle
\raggedbottom

\begin{abstract}
Combining vision-language models (VLMs) with parameter-efficient fine-tuning (PEFT) offers a promising approach to few-shot whole-slide image (WSI) classification in computational pathology. This study evaluates this approach for predicting ethanol drinking status from nonhuman primate liver histology. Adapters are evaluated against non-PEFT configurations on a subset of the Monkey Alcohol Tissue Research Resource (MATRR). All evaluated PEFT configurations achieve higher mean balanced accuracy than the non-PEFT baselines. For example, in the four-shot setting, attention-based multiple instance learning (ABMIL) alone achieves 43.5\% mean balanced accuracy, compared with 78.1\% after adding an adapter. Qualitative attention analysis shows that textual pathway adaptations produce localized attention heatmaps, while visual pathway adaptations produce more diffuse ones. A pathologist’s review identified similar histological features across adaptations.
\end{abstract}


\section{Introduction}

Alcohol-associated liver disease (ALD) is challenging to investigate in human populations due to confounding genetic and lifestyle factors~\cite{mackowiak2024alcohol}. Animal models provide controlled experimental settings for studying alcohol-induced liver pathology and disease progression~\cite{cao2026update}. In particular, nonhuman primates exhibit histological features that resemble human liver disease, making them valuable for translational research~\cite{wang2015}. Previous studies have demonstrated the feasibility of predicting ethanol drinking categories from nonhuman primate liver histology using deep learning~\cite{khanal2025}. However, the application of vision-language models (VLMs) to this setting remains underexplored, particularly when labeled data are limited.

In computational pathology, a typical data point is a gigapixel-resolution whole slide image (WSI), which presents challenges for learning from slide-level labels when disease-related features may be localized to small regions across the slide~\cite{bogahawatte2026parameter}. To address the limitations of relying on exhaustive pixel-level manual annotations, multiple instance learning (MIL) is commonly used. MIL treats each slide as a bag of patches and learns slide-level predictions without patch annotations. Attention-based MIL (ABMIL)~\cite{ilse2018} introduces a trainable gated attention mechanism to weight patches based on their importance and form a slide-level representation. However, MIL architectures can add tens of thousands of trainable parameters, which may increase the risk of overfitting when labeled data are limited. Since obtaining large-scale annotated datasets for specific clinical tasks can be unfeasible, there is a need for models that generalize under few-shot, weakly supervised constraints \cite{srinidhi2021deep}. These challenges motivated few-shot weakly supervised WSI classification (FSWC) \cite{qu2023}. In this setting, a parameter-efficient fine-tuning (PEFT) technique \cite{hu2021lora} is leveraged to adapt VLMs using a minimal number of trainable parameters, while keeping the large-parameter-space encoder backbones frozen.

VLMs such as CONtrastive learning from Captions for Histopathology (CONCH)~\cite{lu2024} are pretrained through pretraining on diverse histological image-text pairs to learn transferable visual representations for downstream pathology tasks. Several FSWC approaches incorporate language information to improve slide-level learning. Two-level Prompt Learning (TOP)~\cite{qu2023} uses pathology language priors to guide patch aggregation and prompt learning, while FOCUS~\cite{guo2025} filters redundant patch features and uses language guidance to prioritize diagnostically relevant regions. MIL-Adapter~\cite{meseguer2026} combines textual ensemble learning with slide aggregation and vision-language adapters. This study builds upon the MIL-Adapter, and adapts its framework to the Monkey Alcohol Tissue Research Resource (MATRR) dataset~\cite{daunais2014} to investigate few-shot, weakly supervised classification of ethanol drinking status from nonhuman primate liver WSIs, establishing a proof of concept for integration of PEFT and VLM in translational ALD research.
\begin{figure}[!t]
\centering
\includegraphics[width=\columnwidth]{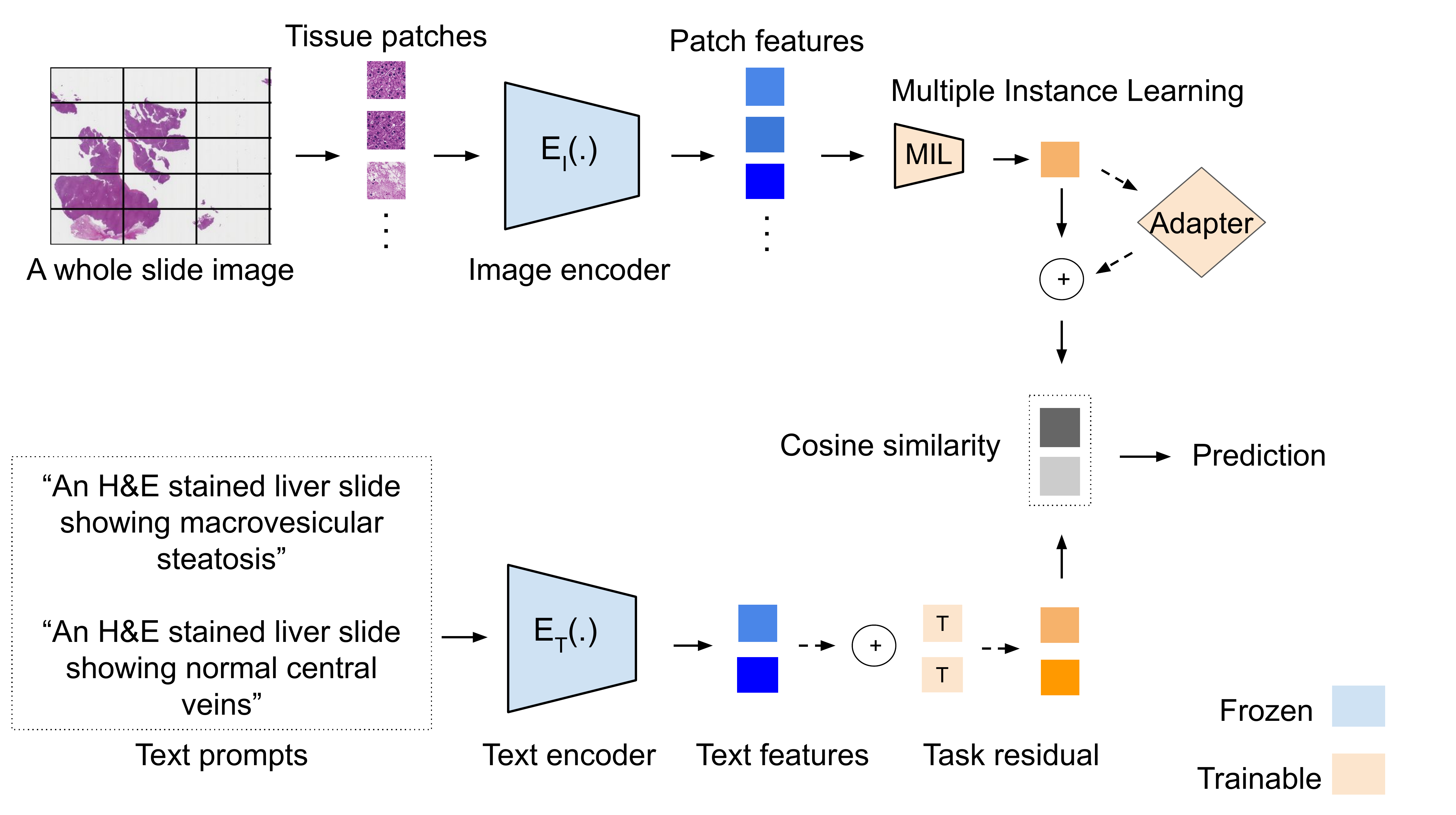}
\caption{Overview of the workflow. The pipeline follows the MIL-Adapter framework~\cite{meseguer2026}, utilizes frozen CONCH encoders~\cite{lu2024} for feature extraction, and is adapted for application to the MATRR dataset~\cite{daunais2014}.}
\label{fig:method}
\end{figure}
\begin{table*}[t]
\centering
\caption{Test set performance (\%, mean $\pm$ standard deviation over 30 runs) on the MATRR dataset across 2-shot and 4-shot scenarios. \textbf{Bold} indicates the highest means.}
\label{tab:results}
\small
\setlength{\tabcolsep}{4pt}
\renewcommand{\arraystretch}{1.08}
\begin{tabular*}{\textwidth}{@{\extracolsep{\fill}}lcccccc@{}}
\toprule
& \multicolumn{3}{c}{2-shot} & \multicolumn{3}{c}{4-shot}\\
\cmidrule(lr){2-4}\cmidrule(l){5-7}
Method & BACC & AUC & F1 score & BACC & AUC & F1 score\\
\midrule
Max Pool & \pmv{35.0}{2.7} & \pmv{19.9}{3.1} & \pmv{57.2}{2.0} & \pmv{36.4}{7.8} & \pmv{19.1}{8.8} & \pmv{63.2}{4.5}\\
ABMIL & \pmv{39.1}{8.6} & \pmv{26.0}{11.5} & \pmv{20.7}{21.2} & \pmv{43.5}{11.8} & \pmv{33.4}{10.4} & \pmv{27.4}{20.0}\\
ABMIL + Adapter $D=128$ & \pmv{52.4}{22.7} & \pmv{51.7}{32.7} & $\mathbf{57.8\!\pm\!22.3}$ & \pmv{76.5}{11.8} & $\mathbf{84.2\!\pm\!12.6}$ & \pmv{78.9}{12.7}\\
ABMIL + Adapter $D=8$ & $\mathbf{57.3\!\pm\!19.6}$ & $\mathbf{56.7\!\pm\!29.4}$ & \pmv{52.3}{28.7} & $\mathbf{78.1\!\pm\!10.8}$ & \pmv{83.6}{11.6} & $\mathbf{83.1\!\pm\!10.0}$\\
ABMIL + TaskRes & \pmv{46.5}{13.4} & \pmv{45.5}{21.9} & \pmv{45.1}{24.6} & \pmv{62.3}{11.7} & \pmv{66.8}{13.1} & \pmv{71.0}{14.2}\\
\bottomrule
\end{tabular*}
\end{table*}
\section{Method}
Figure~\ref{fig:method} provides an overview of the workflow. The framework employs the frozen image and text encoders of CONCH \cite{lu2024}, pre-trained on 1.17 million histopathology image–text pairs. Text prompts are constructed that are relevant to liver pathology. Let $P$ be a set of prefix strings and $M_c$ be a set of morphological descriptors associated with class $c\in\{0,1\}$. The prefixes are ``An H\&E stained liver slide showing'' and ``Histopathological examination of a liver slide displaying''. For the Non-Drinking (Control) class, $M_0$ includes ``orderly trabecular architecture'', ``clear sinusoids'', ``intact portal tracts'', ``normal central veins'', and ``unremarkable parenchyma''. For the Drinking class, $M_1$ includes ``macrovesicular steatosis'', ``hepatocyte ballooning'', ``Mallory-Denk bodies'', ``neutrophilic infiltrate'', and ``pericellular fibrosis''. The Cartesian product of $P$ and $M_c$ generates $K=10$ textual descriptions for each class. The frozen text encoder produces $L_2$-normalized features $\mathbf{T}_{c,k}\in\mathbb{R}^{d}$ for prompt $k$, where $d=512$. Mean pooling over the $K$ features forms the class text vector $\mathbf{t}_c\in\mathbb{R}^{d}$.

On the visual path, tissue is segmented using Otsu thresholding~\cite{otsu1979}. Slides are divided into small patches. The frozen image encoder produces $L_2$-normalized patch features $\mathbf{X}=\{\mathbf{x}_1,\ldots,\mathbf{x}_N\}$, where $N$ is the number of retained patches and $\mathbf{x}_i\in\mathbb{R}^{d}$. ABMIL generates a slide-level representation by learning an attention weight for each patch, as defined in~\eqref{eq:attention}:
\begin{equation}
a_i=\frac{\exp\!\left(\mathbf{w}^{T}[\tanh(\mathbf{V}\mathbf{x}_i)\odot\sigma(\mathbf{U}\mathbf{x}_i)]/\tau\right)}{\sum_{j=1}^{N}\exp\!\left(\mathbf{w}^{T}[\tanh(\mathbf{V}\mathbf{x}_j)\odot\sigma(\mathbf{U}\mathbf{x}_j)]/\tau\right)},
\label{eq:attention}
\end{equation}

where $\mathbf{V},\mathbf{U}\in\mathbb{R}^{64\times 512}$ and $\mathbf{w}\in\mathbb{R}^{64}$ are trainable parameters, $\sigma$ denotes sigmoid, and $\odot$ is the elementwise product. The attention temperature is $\tau=0.33$. The slide-level representation is the weighted sum, as given in~\eqref{eq:slide}:

\begin{equation}
\mathbf{h}_{slide}=\sum_{i=1}^{N}a_i\mathbf{x}_i\in\mathbb{R}^{d}.
\label{eq:slide}
\end{equation}
Cosine similarities between the slide-level and class text representations are multiplied by a learnable scale and passed through softmax to obtain class probabilities. When incorporating PEFT modules, the similarities are computed using the adapted representations. The visual adapter applies a two-layer bottleneck multilayer perceptron (MLP), projecting $\mathbf{h}_{slide}$ to a smaller dimension ($D$) and back, as expressed in~\eqref{eq:adapter}:
\begin{equation}
\mathbf{h}'_{slide}=\beta\,\operatorname{MLP}_{D}(\mathbf{h}_{slide})+(1-\beta)\mathbf{h}_{slide},
\label{eq:adapter}
\end{equation}
where $\beta=0.2$ controls the residual ratio.
\par

Task residuals (TaskRes), on the textual path, instead optimize a learnable vector $\Delta\mathbf{t}_c\in\mathbb{R}^{d}$ for each class, as shown in~\eqref{eq:taskres}:
\begin{equation}
\mathbf{t}'_c=\mathbf{t}_c+\alpha\,\Delta\mathbf{t}_c,
\label{eq:taskres}
\end{equation}
where $\alpha=0.25$ scales the residual. Cross-entropy loss on slide-level labels is used for training.
\section{Results}
The MATRR subset (cohorts 6a, 6b, 7a, 7b) consists of 37 WSIs from distinct rhesus macaques, comprising 13 Non-Drinking (Control) subjects and 24 Drinking subjects. Training and validation each contain two or four subjects per class; the remaining subjects are held out for testing. Each configuration is evaluated over 30 runs (10 iterations across three seeds). Models use AdamW with a base learning rate of 0.001.
Training runs for a maximum of 250 epochs with an early stopping patience of 50 epochs. The checkpoint with the lowest validation loss is selected. Table~\ref{tab:results} reports classification performance as the balanced accuracy (BACC), the area under the receiver operating characteristic curve (AUC), and the F1 score.

All evaluated PEFT configurations achieve higher mean BACC and AUC than both baselines at both shot settings. Relative to ABMIL, the $D=8$ adapter increases mean BACC from 43.5\% to 78.1\% at four shots, while mean AUC increases from 33.4\% to 83.6\%. The visual pathway adapters achieve higher mean performance than TaskRes. The larger standard deviations suggest greater sensitivity to the sampled subjects, consistent with the variability commonly observed in few-shot learning due to limited training examples~\cite{wang2020frustratingly}.

\begin{figure}[!t]
\centering
\includegraphics[width=\columnwidth]{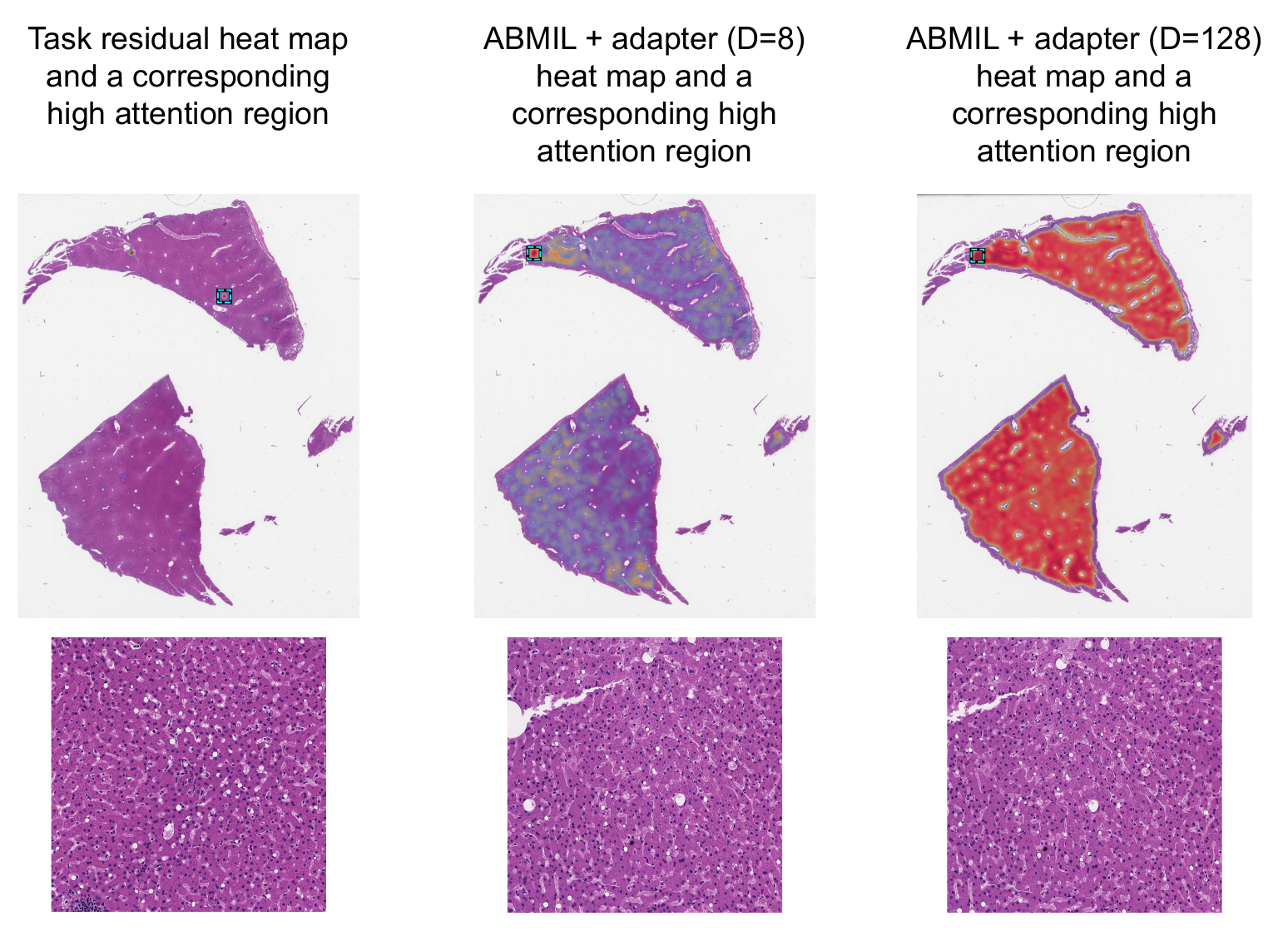}
\caption{Attention heatmaps and corresponding high-attention regions for a Drinking subject in the four-shot setting. TaskRes (left) shows localized attention, while visual adapters with $D=8$ (middle) and $D=128$ (right) attend more broadly.}
\label{fig:xai}
\end{figure}

Figure~\ref{fig:xai} shows ABMIL attention weights mapped to patch coordinates and overlaid on the WSI of a Drinking subject in the four-shot setting. For this slide, TaskRes produces localized attention, whereas the visual adapters show broader attention distributions. Although all configurations learn patch attention through ABMIL, TaskRes adapts the class text features, whereas the visual adapters modify the aggregated slide representation. These differences in architecture and the placement of adaptation modules may contribute to the observed attention patterns.

A pathologist review of the attention heatmaps and callouts in Figure \ref{fig:xai} found general concordance across the three models, with all three highlighting features consistent with rare single-cell necrosis, increased sinusoidal space, and/or minimal Ito cell hyperplasia. The ABMIL Adapter (D = 8) additionally identified a single focus of minimal microvesicular hepatocyte vacuolation or degeneration that was not flagged by the other models. Both ABMIL Adapters (D = 8 and D = 128) highlighted similar features, but the one with a lower bottleneck dimension was judged more discriminating overall. Although further validation is required, this suggests that the approach is a more promising candidate for further development.

\section{Conclusion}

In this work, a proof of concept is established for few-shot ethanol drinking-status classification from nonhuman primate liver histology. The investigation shows that MIL alone struggles to generalize directly from foundation model features, but integrating adaptation modules improves the classification performance in this setting. The illustrated heatmaps show localized attention for textual adapter and more diffuse attention for visual adapters, indicating that the two approaches produce different spatial patterns. Future work will evaluate the framework on larger datasets and test prompt sensitivity.

\ifdoubleblind
\else
\section*{Acknowledgment}
The authors would like to thank Dr. Nathan Crilly for providing the pathology review for this study. The MATRR is supported by the National Institutes of Health (AA019431).
\fi

\bibliographystyle{IEEEtran}
\bibliography{ref}

\end{document}